\documentclass[sigconf]{acmart}

\setcopyright{none}
\renewcommand\footnotetextcopyrightpermission[1]{}

\AtBeginDocument{%
  }

\usepackage{subcaption}
\graphicspath{{graphics/}}

\begin{document}

\title{User-Centric Modeling of Transactional Sequences with Explainable State Space Models}

\author{Ivan Palagin}
\email{idpalagin@edu.hse.ru}
\affiliation{%
  \institution{HSE University}
  \city{Moscow}
  \country{Russia}
}

\renewcommand{\shorttitle}{User-Centric Modeling of Transactional Sequences with Explainable State Space Models}
\renewcommand{\shortauthors}{I.D. Palagin}

\begin{abstract}
We propose a hybrid approach for user-centric modeling of transactional event sequences
that combines contrastive representation learning (CoLES) with State Space Models (SSMs).
While contrastive methods yield high-quality compressed user representations, existing
encoders---RNNs and Transformers---suffer from vanishing gradients or quadratic
complexity, respectively. Mamba, a selective SSM, efficiently handles long-range
dependencies but remains underexplored for personalized user analysis. We investigate
two integration strategies: (1)~initializing the Mamba hidden state with a CoLES
embedding, and (2)~prepending the projected CoLES embedding as a prefix token to the
input sequence. Both approaches supply the model with an informative user prior from the
first step. Experiments on three public datasets---Age (multiclass age-group prediction),
MBD (multi-label product acquisition), and Taobao (binary purchase prediction)---demonstrate
consistent improvements over standalone Mamba and CoLES with a linear classifier, with
the hybrid models converging 2--3$\times$ faster than the plain SSM baseline.
Explainability analysis via discretization-step maps and Integrated Gradients reveals
selective event filtering on behavior-rich datasets and identifies the most informative
transaction features.
\end{abstract}



\maketitle

\makeatletter
\def\acmConference@shortname{HSE University}
\def\acmConference@date{2026}
\def\acmConference@venue{Moscow, Russia}
\makeatother

\section{Introduction}

In modern digital services virtually every user interaction is recorded as a timestamped
event. Sequences of such events---purchases, page views, financial transactions---encode
rich behavioral patterns commonly referred to as \emph{transactional event sequences}.
Accurate modeling of these sequences is critical for fraud detection, credit scoring,
churn prediction, and personalized recommendation~\cite{osin2025ebes,springer_ch51}.

A particularly valuable objective is learning \emph{user-level representations}: a
compact profile that summarizes behavioral history can serve as a powerful prior for
multiple downstream tasks without reprocessing the full sequence at inference time.
Contrastive Learning for Event Sequences (CoLES)~\cite{babaev2022coles} is a prominent
self-supervised paradigm for this purpose, constructing positive pairs through sequence
augmentations (masking, time jitter, within-window permutation) and optimizing a
contrastive objective. The resulting embeddings capture stable user profiles without
label supervision. However, CoLES depends on a sequential encoder backbone, in practice
an RNN or Transformer, both of which have fundamental limitations for long sequences.

Table~\ref{tab:arch_comparison} summarizes the trade-offs among encoder architectures.
RNNs process events strictly sequentially and suffer from vanishing gradients, preventing
reliable long-range dependency capture. Transformers overcome this via self-attention but
at $\mathcal{O}(L^2)$ complexity, which is prohibitive for banking sequences averaging
881 events/user (Age dataset, 21M total events) or 156M total events (MBD). State Space
Models (SSMs)---specifically Mamba~\cite{gu2023mamba}---offer $\mathcal{O}(L)$ complexity,
stable long-range memory via HiPPO initialization~\cite{gu2020hippo}, and a parallelizable
convolutional training form.

\begin{table}[h]
  \caption{Comparison of sequence encoder architectures.}
  \label{tab:arch_comparison}
  \small
  \begin{tabular}{lccc}
    \toprule
    \textbf{Architecture} & \textbf{Long-range} & \textbf{Complexity} & \textbf{Parallel} \\
    \midrule
    GRU / LSTM   & \texttimes & $\mathcal{O}(L)$   & \texttimes \\
    Transformer  & \checkmark & $\mathcal{O}(L^2)$ & \checkmark \\
    Mamba (SSM)  & \checkmark & $\mathcal{O}(L)$   & \checkmark \\
    \bottomrule
  \end{tabular}
\end{table}

Despite these advantages, SSMs remain underexplored in the context of \emph{user-centric}
personalized modeling. This work closes the gap with two hybrid architectures that inject
CoLES user embeddings into Mamba: \textbf{(1)~hidden-state initialization}---the CoLES
embedding is projected into Mamba's initial hidden state $\mathbf{h}_0$; and
\textbf{(2)~prefix concatenation}---the projected embedding is prepended as an extra
context token (Figure~\ref{fig:architectures}). Both strategies are evaluated with
per-pair Optuna~\cite{akiba2019optuna} hyperparameter tuning, and model decisions are
interpreted with discretization-step maps and Integrated Gradients~\cite{sundararajan2017axiomatic}.

\section{Related Work}

\textbf{Event sequence modeling.}
EBES~\cite{osin2025ebes} provides a large-scale benchmark showing that GRU-like
models~\cite{hidasi2016gru4rec} dominate on short sequences while long-range architectures
become competitive as sequence length grows. Transformer-based encoders such as
BST~\cite{chen2019bst} and BERT4Rec~\cite{sun2019bert4rec} achieve strong results on
behavioral sequences from e-commerce platforms but incur quadratic attention overhead
that is prohibitive for banking sequences with hundreds to thousands of events per user.

\textbf{LLM-based and hybrid approaches.}
LLM4ES~\cite{shestov2025llm4es} learns user embeddings by converting event sequences
into textual descriptions fed to a language model, demonstrating strong zero-shot
transfer to downstream tasks. LATTE~\cite{fadeev2025latte} reduces inference costs by
aligning compact transaction embeddings with LLM-generated behavioral summaries via
contrastive training, showing that frozen LLM semantic priors improve downstream quality
at lower latency. EAFD~\cite{sakhno2026eafd} uses a pretrained-embedding-guided LLM
agent to discover interpretable features complementary to the latent representation,
achieving up to $+5.8\%$ relative gains over state-of-the-art embeddings on transaction
benchmarks. FinTRACE~\cite{sakhno2026fintrace} proposes a retrieval-first pipeline that
converts raw transaction histories into behavioral features stored in a knowledge base,
enabling grounded LLM reasoning for financial analytics in low-supervision settings.
Despite competitive accuracy, all LLM-based approaches impose substantial inference
overhead, motivating lightweight SSM-based alternatives~\cite{springer_ch51}.

\textbf{Contrastive learning for event sequences.}
CoLES~\cite{babaev2022coles} pioneered self-supervised contrastive learning for event
sequences. MLEM~\cite{moskvoretskii2024mlem} combines contrastive and generative
objectives; Yugay and Zaytsev~\cite{yugay2025uniting} study joint training strategies,
both showing gains over pure contrastive baselines. Our work is complementary: we keep
CoLES fixed as a pretrained user-profile module and focus on integrating its embeddings
into a more expressive SSM-based downstream encoder.

\textbf{State Space Models.}
S4~\cite{gu2022s4} introduced structured SSMs with near-linear complexity, setting
competitive Long Range Arena~\cite{tay2021lra} scores. Mamba~\cite{gu2023mamba} added
selective state gating via input-dependent $B_t$, $C_t$, $\Delta_t$; Mamba-2~\cite{dao2024mamba2}
simplified the state matrix to scalar-diagonal form, yielding faster training and improved
long-range retention. ReMamba~\cite{yuan2024remamba} explores selective compression for
very long contexts. A comprehensive survey of SSM variants is provided
in~\cite{s4survey2025}. At scale, Mamba-based language models match Transformers on most
tasks while offering up to $8\times$ inference speedup~\cite{waleffe2024mamba_lm}.

\textbf{Explainability of Mamba.}
Ali et al.~\cite{ali2025hidden} proved that selective SSMs are mathematically equivalent
to linear-attention models, enabling derivation of implicit attention matrices and direct
application of gradient-based XAI methods. Yang and Wang~\cite{yang2025evm} proposed
discretization-step maps and cross-modal attention weights for Vision Mamba; these maps
generalize to any sequential domain and constitute the primary interpretability tool
in this work alongside Integrated Gradients~\cite{sundararajan2017axiomatic}.

\section{Methodology}

\subsection{Datasets}

Experiments are conducted on three public benchmarks (Table~\ref{tab:datasets}).
\textbf{Age} contains real banking transactions labeled with one of four age groups
(balanced, 25\% each; evaluated by Accuracy). \textbf{MBD} contains banking transactions
with four binary product-acquisition labels (severe class imbalance: $\approx\!0.3\%$
positive per label; evaluated by mean ROC-AUC). \textbf{Taobao} captures e-commerce
user behavior; the target is whether a user makes a purchase within 7~days (43/57 class
split; evaluated by ROC-AUC).

\begin{table}[h]
  \caption{Dataset characteristics.}
  \label{tab:datasets}
  \begin{tabular}{llll}
    \toprule
    & \textbf{Age} & \textbf{MBD} & \textbf{Taobao} \\
    \midrule
    Task        & Multiclass   & Multi-label  & Binary      \\
    Metric      & Accuracy     & Mean ROC-AUC & ROC-AUC     \\
    Clients     & 24K          & 781K         & 18K         \\
    Events      & 21M          & 156M         & 5.1M        \\
    Avg.\ len   & 881          & 21/window    & 280         \\
    Cat.\ feat  & 1            & 11           & 2           \\
    Num.\ feat  & 1            & 1            & 0           \\
    \bottomrule
  \end{tabular}
\end{table}

\subsection{Feature Preprocessing}

Numerical features are transformed via $\log(1+|x|)\!\cdot\!\text{sign}(x)$, normalized
with per-sequence $z$-score, and projected with a linear layer. Categorical features are
mapped to learned embedding tables. Timestamps are encoded as successive differences
$\delta t_i = t_i - t_{i-1}$ and projected analogously to numerical features. All
per-feature vectors are concatenated to form the per-event input token
$\mathbf{x}_t \in \mathbb{R}^{d_\text{model}}$. The timestamp encoding strategy
(difference, categorical bucketing, or none) is treated as a tunable hyperparameter.

\subsection{Baseline Models}

\textbf{Plain Mamba.} Transaction tokens are projected into the model dimension
$d_\text{model}$, processed by a stack of Mamba blocks (each with RMSNorm and residual
connection), and the last hidden state is fed to a linear classifier. The initial hidden
state $\mathbf{h}_0$ is the zero vector.

\textbf{CoLES + linear classifier.} CoLES is trained self-supervisedly on the full
dataset (train + test union; no label leakage) for 100K gradient steps. User embeddings
are extracted and cached. A linear classifier trained on these embeddings isolates the
quality of the contrastive representation as a standalone predictor.

\subsection{Proposed Approaches}

\begin{figure}[t]
  \centering
  \IfFileExists{graphics/fig_architectures.pdf}{%
    \includegraphics[width=\columnwidth]{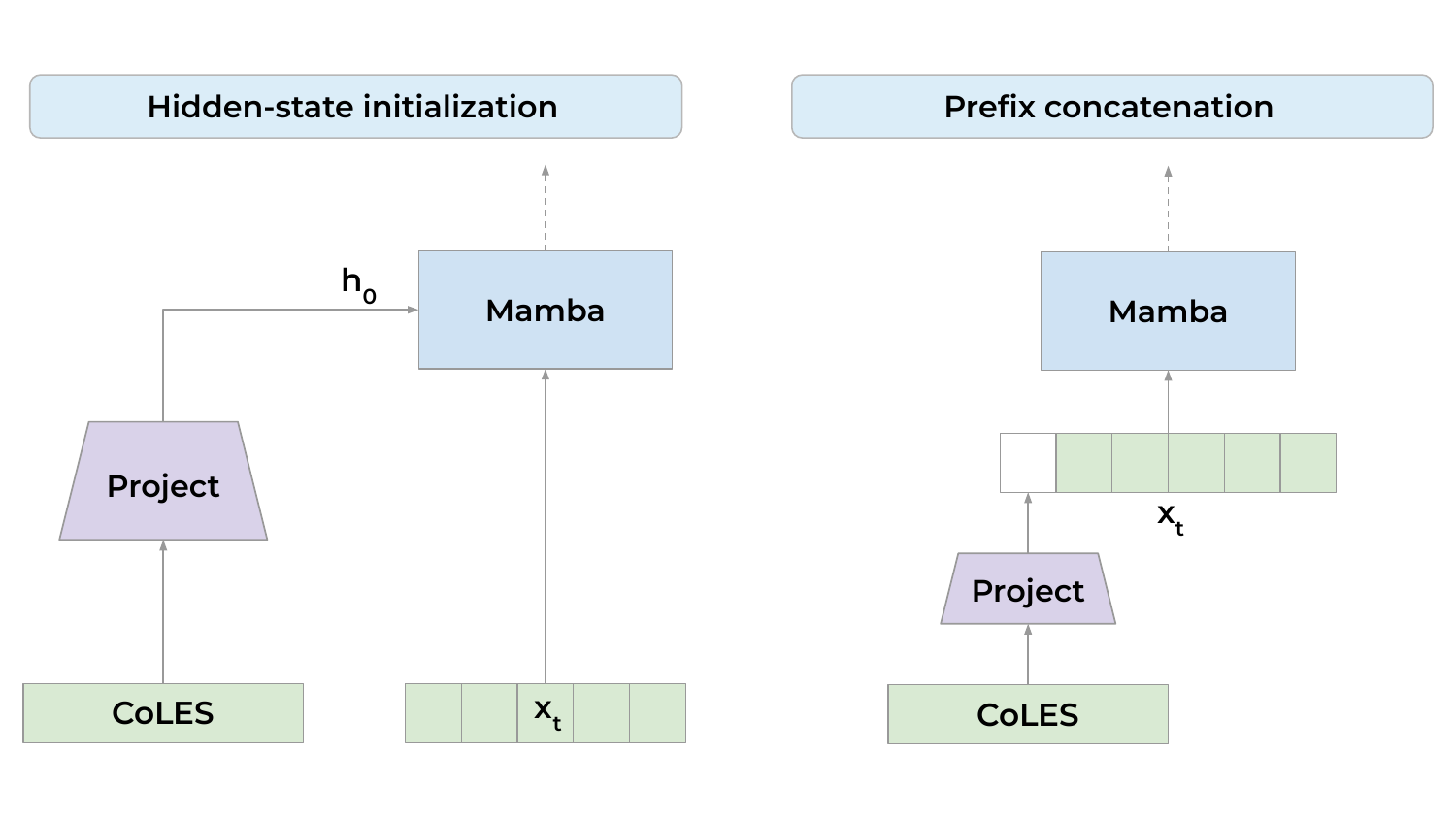}%
  }{%
    \fbox{\parbox{0.9\columnwidth}{\centering\small\vspace{0.8cm}%
      [fig\_architectures.pdf]\vspace{0.8cm}}}%
  }
  \caption{Proposed hybrid architectures. \emph{Left}: hidden-state initialization---the
    CoLES embedding is projected into Mamba's initial state $\mathbf{h}_0$.
    \emph{Right}: prefix concatenation---the embedding is prepended as a context token
    before the transaction sequence $\mathbf{x}_t$.}
  \label{fig:architectures}
\end{figure}

Figure~\ref{fig:architectures} illustrates both proposed hybrid methods. In both cases
CoLES is pretrained independently and its weights are frozen during the downstream phase.

\textbf{Hidden-state initialization.}
We replace the zero initial state with a two-step projection of the user embedding:
\begin{equation}
  \mathbf{h}_0 = W_2\,\sigma(W_1\,\mathbf{e}_\text{CoLES}),
\end{equation}
where $W_1\!\in\!\mathbb{R}^{d_\text{mid}\times d_\text{coles}}$ and
$W_2\!\in\!\mathbb{R}^{(d_N \cdot L_\text{ssm})\times d_\text{mid}}$ are learnable. This
initialization is applied only to the first Mamba layer; subsequent layers receive the
preceding layer's output state. The user profile is thus available as a prior before
the first transaction is processed.

\textbf{Prefix concatenation.}
The CoLES embedding is projected into the transaction token space and prepended to the
input sequence:
\begin{equation}
  \tilde{\mathbf{X}} = [\,P\,\mathbf{e}_\text{CoLES}\ ;\ \mathbf{X}\,],
\end{equation}
where $P\!\in\!\mathbb{R}^{d_\text{model}\times d_\text{coles}}$. Mamba can learn at
each subsequent position how strongly to condition on the user prior in the prefix.

\subsection{Hyperparameter Optimization}

For each (architecture, dataset) pair we run Optuna~\cite{akiba2019optuna} with
Tree-structured Parzen Estimation (TPE) for 50 trials. To keep tuning tractable:
epochs are capped at~10, sequence lengths are halved, and for MBD a 1M-sequence
subsample is used. Key search ranges: learning rate $[10^{-5},\,10^{-1}]$ (log-uniform),
Mamba layers $\{1,2,3\}$, state-space dimension $\{16,32\}$, timestamp encoding
$\{\text{diff},\text{cat},\text{none}\}$, and categorical/numerical embedding dimensions
$[4,64]$. The first 5 trials use random sampling; TPE guides the remaining search.
Results are stored in a SQL database for resumable, reproducible tuning.

\subsection{Training Protocol}

All models are optimized with AdamW. CoLES pretraining runs on the train$\cup$test union
for 100K gradient steps with batch sizes in $[32,\,256]$; embeddings are extracted and
cached for all users. Plain Mamba and both hybrid models are then trained on the training
split with early stopping (patience $= 3$, max 100 epochs) monitored on the held-out
validation metric. For users without a precomputed embedding at inference time, a zero
vector is substituted, reducing the hybrid model exactly to plain Mamba and providing a
natural cold-start fallback.

\section{Results}

\subsection{Main Results}

Table~\ref{tab:results} reports test-set performance. Both hybrid methods consistently
outperform all baselines on all three datasets. The largest absolute gains appear on
Age ($+3.2$\,pp for hidden-state initialization over plain Mamba) and MBD ($+2.6$\,pp).
On Taobao improvements are smaller ($+0.7$\,pp) but consistent across both methods.
CoLES with a linear classifier underperforms plain Mamba on all three datasets, confirming
that sequential context provided by Mamba is important and that CoLES embeddings are most
effective as a complementary user-level prior rather than a standalone predictor.

\begin{table}[h]
  \caption{Test-set performance. \textbf{Bold}: best per dataset.}
  \label{tab:results}
  \begin{tabular}{llll}
    \toprule
    \textbf{Model} & \textbf{Age} & \textbf{MBD} & \textbf{Taobao} \\
                   & Accuracy     & mROC-AUC     & ROC-AUC \\
    \midrule
    Mamba              & 0.354          & 0.711          & 0.692          \\
    CoLES + linear     & 0.321          & 0.703          & 0.641          \\
    Mamba + init       & \textbf{0.386} & \textbf{0.737} & 0.695          \\
    Mamba + prefix     & 0.384          & 0.727          & \textbf{0.699} \\
    \bottomrule
  \end{tabular}
\end{table}

\subsection{Convergence Analysis}

Learning-curve analysis reveals a second advantage of the hybrid architectures beyond
final metric values. Both CoLES-augmented models reach peak validation performance at
epochs~2--3, while plain Mamba requires~6 epochs to converge. This accelerated convergence
indicates that the user prior injected via CoLES---whether as an initial hidden state or
as a prefix token---provides the model with a meaningful starting point that reduces the
number of gradient steps needed to learn user-level behavioral patterns from the raw
transaction stream. Faster convergence is practically relevant for large-scale industrial
deployments where training time and compute costs are critical constraints.

\subsection{Hyperparameter Sensitivity}

Tuning across all (model, dataset) pairs reveals stable patterns: (i)~optimal learning
rates cluster around $10^{-4}$; higher values cause training instability for all
architectures; (ii)~three Mamba layers consistently outperform shallower configurations;
(iii)~categorical and numerical embedding dimensions in $[15,40]$ balance expressiveness
and parameter cost; (iv)~for long sequences (Age, avg.~881 events), a state-space
dimension of~32 outperforms~16, while short-sequence datasets (MBD: 21 events/window)
show less sensitivity to this parameter.

\subsection{Explainability Analysis}

\begin{figure}[t]
  \centering
  \IfFileExists{graphics/fig_explainability.pdf}{%
    \includegraphics[width=\columnwidth]{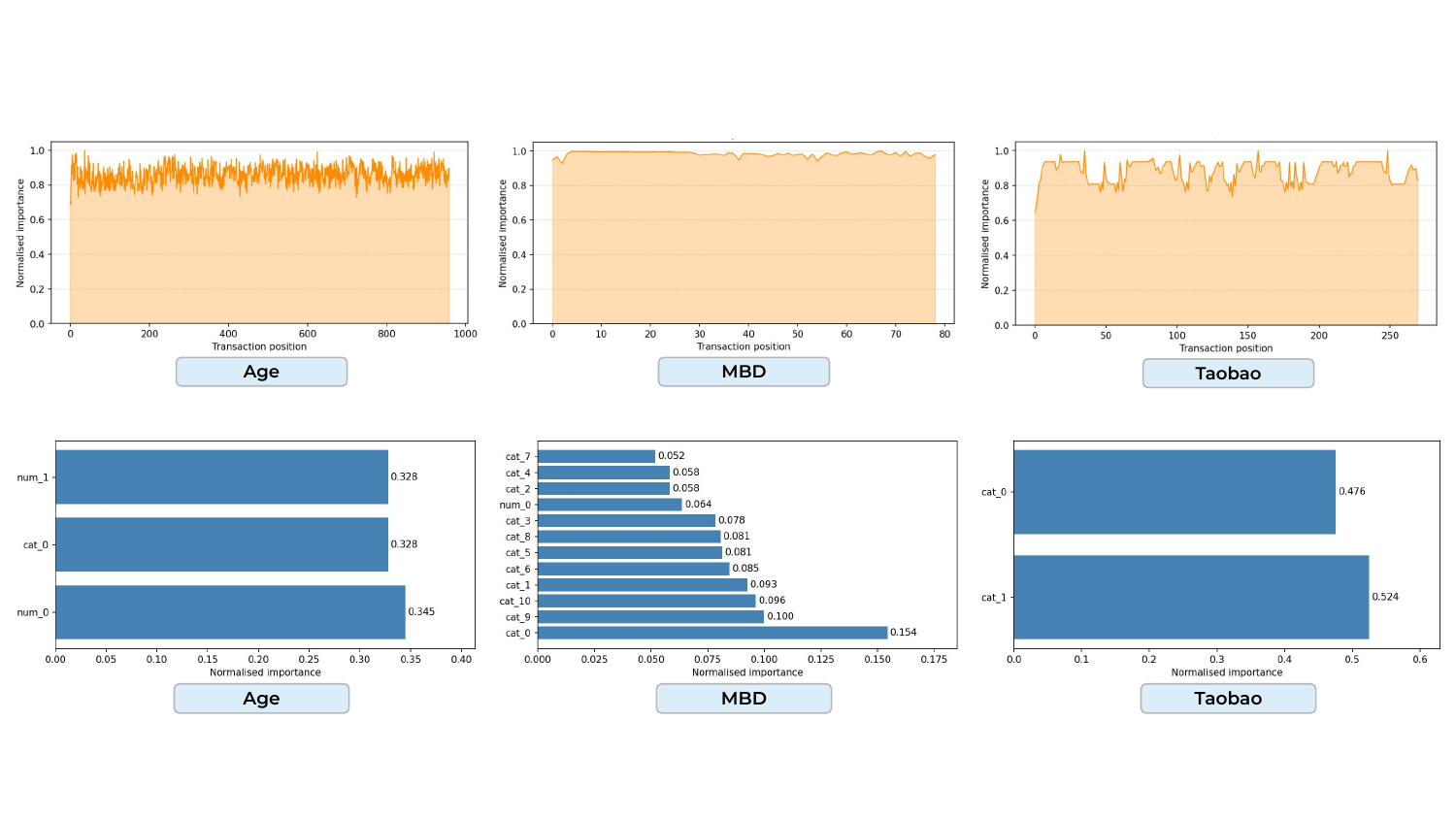}%
  }{%
    \fbox{\parbox{0.9\columnwidth}{\centering\small\vspace{0.8cm}%
      [fig\_explainability.pdf]\vspace{0.8cm}}}%
  }
  \caption{Explainability results for Mamba with hidden-state initialization (16 test
    clients per dataset). \emph{Top row}: normalized discretization-step ($\Delta_t$)
    curves (Age, Taobao, MBD). Pronounced dips in Taobao indicate selective event
    filtering. \emph{Bottom row}: feature importance from Integrated Gradients. Currency
    (\texttt{cat\_0}) dominates in MBD; importance is near-uniform in Age and Taobao.}
  \label{fig:explainability}
\end{figure}

We apply two complementary XAI techniques to the hidden-state initialization model,
sampling 16 test clients per dataset. Figure~\ref{fig:explainability} summarizes the
results.

\textbf{Discretization-step maps.}
The scalar $\Delta_t$ controls how strongly the hidden state is updated at position~$t$.
On Age and MBD, $\Delta_t$ lies uniformly in $[0.75,\,1.0]$ across all positions,
indicating that the model updates its state equally on every transaction without selective
filtering. On Taobao, $\Delta_t$ exhibits pronounced dips at specific positions: the
model learns to suppress low-intent events (item browsing) and amplify high-intent ones
(add-to-cart, purchase), reflecting the qualitatively distinct interaction types present
in e-commerce data.

\textbf{Integrated Gradients.}
IG~\cite{sundararajan2017axiomatic} attributes output logits to input features via
path-integral gradients from a zero baseline. On Age, importance is nearly uniform
across transaction amount (\texttt{num\_0}$\approx$0.345), timestamp (\texttt{num\_1}$\approx$0.328),
and category (\texttt{cat\_0}$\approx$0.328). On MBD, the currency feature (\texttt{cat\_0})
dominates with importance~0.154, substantially exceeding all other features, implying
that foreign-currency transactions correlate strongly with holding specific bank products.
On Taobao, importance is split between item identifier (\texttt{cat\_1}$\approx$0.524)
and behavior type (\texttt{cat\_0}$\approx$0.476).

Taken together, selective event filtering emerges only when transaction types carry
qualitatively different behavioral signals (Taobao), while banking datasets elicit
broad, uniform reliance on all transaction features---consistent with the less structured
nature of financial event streams compared to e-commerce interaction logs.

\section{Conclusion}

We presented two hybrid user-centric architectures that integrate frozen CoLES user
embeddings into the Mamba SSM via hidden-state initialization and prefix concatenation.
Experiments on three diverse transactional datasets demonstrate consistent improvements
over standalone Mamba and CoLES baselines, with notably faster convergence (peak
validation at epochs~2--3 vs.~6 for plain Mamba) attributable to the user-level prior.
Explainability analysis confirms that the models develop selective event filtering on
e-commerce data while processing banking sequences more uniformly---findings that are
interpretable in terms of domain knowledge about financial vs.\ behavioral event types.

The moderate magnitude of metric gains suggests that CoLES and Mamba extract partially
overlapping information from the same sequence. Contributing factors include: (1)~CoLES
trains on short random sub-sequences and ignores global event chronology; (2)~the linear
projection from CoLES space into Mamba's state space may be insufficiently expressive;
(3)~separate optimization of CoLES and Mamba misses potential synergies from joint training.

Promising directions for future work include: (i)~joint end-to-end training of CoLES and
Mamba so the contrastive representation adapts to the SSM encoder's needs; (ii)~replacing
the linear projection with a multi-layer or cross-attention module; (iii)~evaluation with
Mamba-2~\cite{dao2024mamba2}, which offers improved long-range retention at lower cost;
(iv)~scaling to sequences with $L\!>\!1000$ events to fully leverage the $\mathcal{O}(L)$
complexity advantage over Transformers; and (v)~combining CoLES-initialized SSMs with
LLM-based pipelines~\cite{sakhno2026eafd,fadeev2025latte} to unite user-level priors with
language grounding and interpretable feature discovery.

\begin{acks}
The author thanks M.\,O.~Makarenko for supervision
and valuable guidance throughout this research.
\end{acks}

\bibliographystyle{ACM-Reference-Format}
\bibliography{paper}

\end{document}